\documentclass[11pt,a4paper]{article}
\usepackage[hyperref]{acl2017}
\usepackage{times}
\usepackage{latexsym}

\usepackage{url}

\aclfinalcopy 

\usepackage{scrextend} 
\usepackage{color,floatrow}
\usepackage{booktabs}
\usepackage{lingmacros}
\usepackage{xspace}
\usepackage{graphicx}
\usepackage{adjustbox}
\usepackage{bbding} 
\usepackage{array}
\usepackage{multirow}

\newcommand{\erstdt}{En-DT}
\newcommand{\enrst}{En-DT}
\newcommand{\pcc}{De-DT}
\newcommand{\derst}{De-DT}

\newcommand{\brrst}{Pt-DT}
\newcommand{\ptrst}{Pt-DT}

\newcommand{\esrst}{Es-DT}
\newcommand{\sprst}{Es-DT}
\newcommand{\durst}{Nl-DT}
\newcommand{\nlrst}{Nl-DT}

\newcommand{\sfudt}{En-SFU-DT}
\newcommand{\instrdt}{En-Instr-DT}
\newcommand{\sfu}{En-SFU-DT}
\newcommand{\instr}{En-Instr-DT}
\newcommand{\gum}{En-Gum-DT}

\newcommand{\rel}[1]{{\sc #1}\xspace}

\usepackage{fancyvrb}

\usepackage{tikz}
\usetikzlibrary{shapes.geometric}
\usetikzlibrary{positioning}
\usepackage{tikz-qtree,tikz-qtree-compat}
\tikzset{every tree node/.style={align=center, anchor=north}}
\usetikzlibrary{arrows}
\newfloatcommand{capbtabbox}{table}[][\FBwidth]

\title{Cross-lingual and cross-domain discourse segmentation \\of entire documents}

\author{Chlo{\'e} Braud \\
  CoAStaL DIKU\\University of Copenhagen\\ University Park 5, \\2100 Copenhagen \\
  {\tt chloe.braud@gmail.com} \\\And
  Oph{\'e}lie Lacroix \\
  CoAStaL DIKU\\University of Copenhagen\\ University Park 5, \\2100 Copenhagen \\
  {\tt lacroix@di.ku.dk} \\\And
  Anders S{\o}gaard \\
  CoAStaL DIKU\\University of Copenhagen\\ University Park 5, \\2100 Copenhagen \\
  {\tt soegaard@di.ku.dk} \\}

\date{}

\begin{document}

\maketitle

\begin{abstract}
  Discourse segmentation is a crucial step in building end-to-end discourse parsers.
However, discourse segmenters only exist for a few languages and domains. Typically they only  detect intra-sentential segment boundaries, assuming gold standard sentence and token segmentation, and relying on high-quality syntactic parses and rich heuristics that are not generally available across languages and domains. In this paper, we propose statistical discourse segmenters for five languages and three domains that do not rely on gold pre-annotations. 
We also consider the problem of learning discourse segmenters when no labeled data is available for a language. 
  Our fully supervised system obtains $89.5$\% F$_1$ for English newswire, with slight drops in performance on other domains, and we report supervised and unsupervised (cross-lingual) results for five languages in total.
\end{abstract}

\section{Introduction}
\label{sec:introduction}

Discourse segmentation is the first step in building a discourse parser.
The goal is to identify the minimal units --- called Elementary Discourse Units (EDU) --- in the documents that will then be linked by discourse relations. 
For example, the sentences (\ref{ex:but}) and (\ref{ex:said})\footnote{The
examples come from the RST Discourse Treebank.} are each segmented into two
EDUs, then respectively linked by a \rel{Contrast} and an \rel{Attribution}
relation.
The EDUs are mostly clauses and may cover a full sentence.
This step is crucial: 
making a segmentation error leads to an error in the final analysis.
Discourse segmentation can also inform other tasks, such as argumentation mining, anaphora resolution, or speech act assignment~\cite{sidarenka:discourse:2015}.

\eenumsentence{
\item $[$Such trappings suggest a glorious past$]$ $[$but give no hint of a troubled present.$]$
\label{ex:but}
\item $[$He said$]$ $[$the thrift will to get regulators to reverse the decision.$]$
\label{ex:said}
}

We focus on the Rhetorical Structure Theory (RST)  \cite{mann:rhetorical:1988} -- and resources such as the RST Discourse Treebank (RST-DT) \cite{carlson:building:2001} -- in which discourse structures are trees covering the documents.
Most recent works on RST discourse parsing focuses on the task of tree building, relying on a gold  discourse segmentation 
\cite{ji:representation:2014,feng:linear:2014,li:recursive:2014,joty:combining:2013}.
However, discourse parsers' performance drops by $12$-$14$\% when relying on \emph{predicted} 
segmentation~\cite{joty:codra:2015}, underscoring the importance of discourse segmentation.
State-of-the-art performance for discourse segmentation on the RST-DT is about
$91$\% in F$_1$ with predicted parses~\cite{xuanbach-leminh-shimazu:2012}, but these  systems rely on a gold segmentation of sentences and words, therefore probably overestimating performance \textit{in the wild}.  
We propose to build discourse segmenters without making any data assumptions. 
Specifically, rather than segmenting sentences, our systems segment documents directly.

Furthermore, only a few systems have been developed for languages other than English and domains other than the Wall Street Journal texts from the RST-DT. 
We are the first to perform experiments across 5 languages, and 3 non-newswire English domains.
Since our goal is to provide a system usable for low-resource languages, we only use language-independent resources: here, the Universal Dependencies (UD) \cite{ud13} Part-of-Speech (POS) tags, for which annotations exist for about 50 languages. 
For the cross-lingual experiments, we also rely on cross-lingual word embeddings induced from parallel data. With a shared representation, we can transfer model parameters across languages, or learn models jointly through multi-task learning. 

\textbf{Contributions: } We (i) propose a general statistical discourse segmenter (ii) that does not assume gold sentences and tokens, and (iii) evaluate it across 5 languages and 3 domains.

We make our code available at \url{https://bitbucket.org/chloebt/discourse}.

\section{Related work}
\label{sec:related}

For English RST-DT, the best discourse segmentation results were presented in~\newcite{xuanbach-leminh-shimazu:2012} ($F_1$ $91.0$\% with automatic parse, $93.7$ with gold parse) -- and in~\newcite{joty:codra:2015} for the Instructional corpus \cite{subba:effective:2009} ($F_1$ $80.9$\% on 10-fold).
Segmenters based on handwritten rules have been developed for Brazilian Portuguese~\cite{pardo:development:2008} ($51.3$\% to $56.8$\%, depending on the genre), Spanish~\cite{dacunha:diseg:2010,cunha:diseg:2012} ($80$\%) and Dutch~\cite{vliet:syntax:2010} ($73$\% with automatic parse, $82$\% with gold parse).\footnote{
For German~\cite{sidarenka:discourse:2015} propose a segmenter in clauses (that may be EDU or not).}

Most statistical discourse segmenters are based on classifiers~\cite{fisher:utility:2007,joty:codra:2015}.
\newcite{subba:automatic:2007} were the first to use a neural network, and 
\newcite{sporleder:discourse:2005}~to model the task as a sequence prediction problem. 
In this work, we do sequence prediction using a neural network.

All these systems rely on a quite large range of lexical and syntactic features (e.g. token, POS tags, lexicalized production rules). 
\newcite{sporleder:discourse:2005}~present arguments for a knowledge-lean system that can be used for low-resourced languages. 
Their system, however, still relies on several tools and gold annotations (e.g. POS tagger, chunker, list of connectives, gold sentences). 
In contrast, we present what is to the best of our knowledge the first work on discourse segmentation that is directly applicable to low-resource languages, presenting results for scenarios where no labeled data is available for the target language. 

Previous work, relying on gold sentence boundaries, also only considers intra-sentential segment boundaries. We move to processing entire documents, motivated by the fact that sentence boundaries are not easily detected across all languages.

\section{Discourse segmentation}
\label{sec:cross}

\paragraph{Nature of the EDUs}

Discourse segmentation is the first step in annotating a discourse corpus.
The annotation guidelines define what is the nature of the EDUs, broadly relying on lexical and syntactic clues.
If sentences and independent clauses are always minimal units, some fine distinctions make the task difficult.

In the English RST-DT \cite{carlson:discourse:2001}, lexical information is crucial: for instance, the presence of the discourse connective ``but" in example (\ref{ex:but})\footnote{All the examples given come from \cite{carlson:building:2001}.} indicates the beginning of an EDU.
In addition, clausal complements of verbs are generally not treated as EDUs. 
Exceptions are the complements of attribution verbs, as in (\ref{ex:said}), and the infinitival clauses marking a \rel{Purpose} relation as the second EDU in (\ref{ex:purpose}).
Note that, in this latter example, the first infinitival clause (``\textbf{to} cover up $\ldots$") is, however, not considered as an EDU.
This fine distinction corresponds to one of the main difficulties of the task.
Another one is linked to coordination: coordinated clauses are generally segmented as in (\ref{ex:coordclause}), but not coordinated verb phrases as in (\ref{ex:coordvp}).

\eenumsentence{
\item $[$A grand jury has been investigating whether officials at Southern Co. accounting conspired \textbf{to} cover up their accounting for spare parts$]$ $[$\textbf{to} evade federal income taxes.$]$
\label{ex:purpose}

\item $[$they parcel out money$]$ $[$so that their clients can find temporary living quarters,$]$ $[$buy food$]~(\ldots)$ $[$and replaster walls.$]$ 
\label{ex:coordclause}

\item $[$Under Superfund, those$]$ $[$who owned, generated or transported hazardous waste$]$ $[$are liable for its cleanup, ($\ldots$)$]$
\label{ex:coordvp}
}

Finally, in a multi-lingual and multi-domain setting, note that all the corpora do not follow the same rules: for example, the relation \rel{Attribution} is only annotated in the English RST-DT and the corpora for Brazilian Portuguese, consequently, complements of attribution verbs are not segmented in the other corpora.

\paragraph{Binary task} As in previous studies, we view segmentation as a binary task at the word level: a word is either an EDU boundary
(label B, beginning an EDU) or not (label I, inside an EDU).
This design choice is motivated by the fact that, in RST corpora, the EDUs cover the documents entirely, and that EDUs mostly are adjacent spans of text.
An exception is when embedded EDUs break up another EDU, as in Example~(\ref{ex:embedded}).
The units $1$ and $3$ form in fact one EDU.
 We follow previous work on treating this as three segments, but note that this may not be the optimal solution. 

\enumsentence{
$[$But maintaining the key components ($\ldots$)$]_1$ $[$-- {\small a stable exchange rate and high levels of imports}~--$]_2$ $[$will consume enormous amounts ($\ldots$).$]_3$~
\label{ex:embedded}
}
\paragraph{Document-level segmentation} 

Contrary to previous studies, we do not assume gold sentences:
Since sentence boundaries are EDU boundaries, our system jointly predicts sentence and intra-sentential EDU boundaries. 

\section{Cross-lingual/-domain segmentation}

Data is scarce for discourse.
In order to build statistical segmenters for new, low-resourced languages and domains,
we propose to combine corpora within a multi-task learning setting (Section~\ref{sec:model}) leveraging data from well-resourced languages or domains. 
Models are trained on several (source) languages (resp. domains) -- each viewed as an auxiliary task -- for building a system for a (target) language (resp. domain).

\paragraph{Cross-domain}
For cross-domain experiments, the models are trained on all the other (source) domains and parameters are tuned on data for the target domain. 
This allows us to improve performance when only few data points (i.e.
development set) are annotated for a specific domain (semi-supervised setting).

\paragraph{Cross-lingual}
For cross-lingual experiments, we tune our system's parameters by training a system on the data for three languages with sufficient amounts of data (namely, German, Spanish and Brazilian Portuguese), and using English data as a development set. 
We then train a new model also using multi-task learning (with these tuned parameters) using only source training data, and report performance on the target test set. 
This allows us to estimate performance when no data is available for the
language of interest (unsupervised adaptation).

\section{Multi-task learning}
\label{sec:model}

Our models perform sequence labeling based on a stacked $k$-layer bi-directional LSTM, a variant of LSTMs \cite{Hochreiter:Schmidhuber:97} that reads the input in both regular and reversed order, allowing to take into account both left and right contexts \cite{graves:framewisephoneme:2005}. For our task, this enables us, for example, to distinguish between coordinated nouns and clauses.
This model takes as input a sequence of words (and, here, POS tags) represented by vectors (initialized randomly or, for words, using pre-trained embedding vectors).
The sequence goes through an embedding layer, and we compute the predictions of the forward and backward states for the $k$ stacked layers.
At the upper level, we compute the softmax predictions for each word based on a linear transformation.
We use a logistic loss.

We also investigate joint training of multiple languages and domains for discourse segmentation.
We thus try to leverage languages and domains regularities by sharing the architecture and parameters through multi-task training, where an auxiliary task is a source language (resp. domain) different from the target language (resp. domain) of interest.
Specifically, we train models based on hard parameters sharing~\cite{Caruana:93,Collobert:ea:11,klerke:improving:2016,plank:multilingual:2016}:\footnote{We used a modified version of \cite{plank:multilingual:2016} fixing the random seed and using standard SGD.} each task is associated with a specific output layer, whereas the inner layers -- the stacked LSTMs -- are shared across the tasks. At training time, we randomly sample data points from one task and do forward predictions. During backpropagation, we modify the weights of the shared layers and the task-specific outer layer. The model is optimized for one target task (corresponding to the development data used). Except for the outer layer, the target task model is thus regularized by the induction of auxiliary models.

\section{Corpora}
\label{sec:data}

Table~\ref{table:stats} summarizes statistics about the data. 
For English, we use four corpora, allowing us to evaluate cross-domain performance: the RST-DT (\erstdt) composed of Wall Street Journal articles; 
the SFU review corpus\footnote{\url{https://www.sfu.ca/~mtaboada}} (\sfudt) containing product reviews;  
the instructional corpus (\instrdt)~\cite{subba:effective:2009} built on instruction manuals; and the GUM corpus\footnote{\url{https://corpling.uis.georgetown.edu/gum/}} (\gum) containing interviews, news, travel guides and how-tos.

\begin{table}[t!]
\resizebox{\textwidth}{!}{
\begin{tabular}{l|rr|rr}
\toprule
Corpus              & \#Doc      & \#EDU    & \#Sent 	& \#Words          	\\
\midrule
\sfu 				& $400$		& $28,260$	& $16,827$	& $328,362$			\\
\erstdt 			& $385$     & $21,789$ 	& $9,074$ 	& $210,584$      	\\
\ptrst 				& $330$ 	& $12,594$  & $4,385$ 	& $136,346$			\\
\sprst
                    & $266$     & $3,325$   & $1,816$    & $57,768$          \\
\instr 				& $176$		& $5,754$	& $3,090$	& $56,197$			\\
\derst 				& $174$     & $2,979$   & $1,805$    & $33,591$          \\
\midrule
\gum 				& $54$ 		& $3,151$	& $2,400$	& $44,577$			\\
\nlrst              & $80$      & $2,345$   & $1,692$  	& $25,095$          \\
\end{tabular}}%
 \caption{Number of documents, EDUs, sentences and words (according to UDPipe, see Section \ref{sec:setting}). }
\label{table:stats}
\end{table}

For cross-lingual experiments, we use annotated corpora for Spanish (\sprst)~\cite{dacunha:spanish:2011},\footnote{We only use the test set from the annotator A.} German (\pcc)~\cite{stede:postdam:2004,stede:postdam:2014}, Dutch (\durst)~\cite{vliet:building:2011,redeker:multi:2012} and, for Brazilian Portuguese, we merged four corpora (\brrst)~\cite{cardoso:cstnews:2011,collovini:summit:2007,pardo:rhetalho:2005,pardo:construccao:2003,pardo:relaccoes:2004} as done in~\cite{maziero:adaptation:2015}.

Three other RST corpora exist, but we were not able to obtain cross-lingual word embeddings for Basque~\cite{iruskieta:basque:2013} and Chinese~\cite{wu:new:2016}, and could not obtain the data for Tamil~\cite{subalalitha:approach:2012}.

\section{Experiments}
\label{sec:setting}

\paragraph{Data} We use the official test sets for the \enrst\ ($38$ documents) and the \esrst\ ($84$). For the others, we randomly choose $38$ documents as test set, and either keep the rest as development set (\nlrst) or split it into a train and a development set.

\paragraph{Baselines} 
 As baselines at the document level, we report the scores obtained (a) when only considering the sentence boundaries predicted using UDPipe~\cite{udpipe:2016} (UDP-S),\footnote{\url{http://ufal.mff.cuni.cz/udpipe}} and (b) when EDU boundaries are added after each token PoS-tagged with ``PUNCT" (UDP-P), marking either an inter- or an intra-sentential boundary. 

\paragraph{Systems} As described in Section~\ref{sec:cross}, our systems are either mono-lingual or mono-domain (mono), or based on a joint training across languages or domains (cross). The ``mono" systems are built for the languages and domains represented by enough data (upper part of Table~\ref{table:stats}). The ``cross" models are trained using multi-task learning.

\paragraph{Parameters}
The hyper-parameters are tuned on the development set: number of iterations $i \in \{10,20,30\}$, Gaussian noise $\sigma \in \{0.1, 0.2\}$, and number of dimensions $d \in \{50,500\}$. We fix the number $n$ of stacked hidden layers to $2$ and the size of the hidden layers $h$ to $100$ after experimenting on the \erstdt.\footnote{With $n \in \{1,2,3\}$ and $h \in \{100,200,400\}$).}
Our final models use $\sigma=0.2$ and $d=500$.

\paragraph{Representation} 
We use tokens and POS tags as input data.\footnote{A document is a sequence alternating words and POS. The tokens are labeled with a B or an I, the POS, always labeled with an I, are inserted after each token they refer to.} The aim is to build a representation considering the current word and its context, i.e. its POS and the surrounding words/POS. 
We use the pre-trained UDPipe models to postag the documents for all languages.
We experiment with randomly initialized and pre-trained cross-lingual word embeddings built on Europarl \cite{levy:strong:2017}, keeping either the full $500$ dimensions, or the first $50$ ones. 

\paragraph{Results}

\begin{table}[t!]
\centering
\resizebox{\textwidth}{!}{
\begin{tabular}{cl|c@{~~~~}|c@{~~~~}cc}
\toprule
    & 	 & Mono & Cross & UDP-S & UDP-P \\
\midrule
\parbox[t]{2mm}{\multirow{5}{*}{\rotatebox[origin=c]{90}{languages}}} 
    & \erstdt  & $89.5$ & {\bf 62.4}    & $55.6$     & $57.5$   \\
    & \ptrst &  $82.2$ & {\bf 64.0}     & $49.0$     & $62.5$   \\
    & \esrst  & $79.3$ & 64.3        & {\bf 64.9} & $53.3$   \\
    & \derst  & $85.1$ & {\bf 76.6}     & $69.7$     & $68.7$   \\
    & \nlrst  &  -& {\bf 82.6}          & $80.2$     & $76.6$   \\
\midrule
\parbox[t]{2mm}{\multirow{4}{*}{\rotatebox[origin=c]{90}{domains}}} 
    & \erstdt\ (news) &  $89.5$ & {\bf 63.0} & $55.6$ & $57.5$   \\
    & \sfu      & $85.5$ & {\bf 81.5}       &  $70.2$       & $66.1$\\
    & \instr    & $87.1$ & {\bf 77.7}     &  $66.5$       & $69.5$\\
    & \gum      & - & $68.1$                 &  {\bf 77.2}   & $61.8$\\
\end{tabular}}%
\caption{Results ($F_1$), comparing cross-lingual and cross-domain results with UDPipe. 
}
\label{tab:results}
\end{table}

Our systems are evaluated using $F_1$ over the boundaries (B labels), disregarding the first word of each document. 
Our scores are summarized in Table~\ref{tab:results}. 

Our supervised, monolingual systems unsurprisingly give the best performance, with $F_1$ above $80$\%. 
The results are generally linked to the size of the corpora, the larger the better. Only exception is the \sfu, which, however, include more varied annotation (the authors stated that the annotations ``have not been checked for reliability").

The (semi-supervised) cross-domain setting allows us to present the scores one can expect when only $25$ documents are annotated for a new domain (i.e. the development set for the target domain), and to give the first results on the \gum, but here, our model is actually outperformed by the sentence-based baseline (UDP-S).

The (unsupervised) cross-lingual models are generally largely better than UDPipe. 
These are scores that one can expect when doing cross-lingual transfer to build a discourse segmenter for a new language for which no annotated data are available. 
The performance is still quite high, demonstrating the coherence between the annotation schemes, and the potential of cross-lingual transfer. 
We acknowledge that this is a small set of relatively similar Indo-European languages, however. 

Note that the sentence-based baseline has a high precision (e.g. $96.6$ on \esrst\ against $59.8$ for the cross-lingual system), but a much lower recall, since it mainly predicts the sentence boundaries. 
On corpora that mostly contain sentential EDUs (e.g. \nlrst, see Table~\ref{table:stats}), this is a good strategy.
Using the punctuation (UDP-P) could be a better approximation for corpora with more varied EDUs, see the large gain for the \ptrst\ and the \instr.

Our scores are not directly comparable with sentence-level state-of-the-art systems (see Section \ref{sec:related}).
However, for \erstdt, our best system correctly identifies $950$ sentence boundaries out of $991$, but gets only $84.5$\% in $F_1$ for intra-sentential boundaries,\footnote{This score ignores the sentences containing only one EDU~\cite{sporleder:discourse:2005}.} thus lower than the state-of-the-art ($91.0$\%). 
This is because we consider much less information, and because the system was not optimized for this task. 
Interestingly, our simple system beats HILDA \cite{duverle:hilda:2010} (74.1\% in $F_1$), is as good as the other neural network based system \cite{subba:automatic:2007}, and is close to SPADE \cite{soricut:sentence:2003} (85.2\% in $F_1$) \cite{joty:codra:2015}, while all of these systems use parse tree information.

Finally, looking at the errors of our system on the \erstdt, we found that most
of them are on the tokens ``to" (30 out of 94 not predicted as 'B') and ``and"
(24 out of 103), as expected given the annotation guidelines (see Section
\ref{sec:cross}).
These words are highly ambiguous regarding discourse segmentation (e.g. in the test set, $42.3$\% of ``and" indicates a boundary).
We also found errors with coordinated verb phrases -- e.g. ``$[$when rates are rising$]$ 
    $[$\textbf{and} shift out at times$]$" -- that should be split \cite{carlson:building:2001}, a distinction hard to make without syntactic trees.
Finally, since we use predicted POS tags, our system learns from noisy data and makes errors due to postagging and tokenisation errors.

\section{Conclusion}
We proposed new discourse segmenters with good performance for many languages and domains, at the document level, within a fully predicted setting and using only language independent tools. 

\section*{Acknowledgements}

We would like to thank the anonymous reviewers for their comments.
This research is funded by the ERC Starting Grant LOWLANDS No. 313695.

\bibliography{discourse}
\bibliographystyle{acl_natbib}

\end{document}